\documentclass[letterpaper, 10 pt, journal]{IEEEtran}
\IEEEoverridecommandlockouts                              

\usepackage{soul} 

\usepackage[usenames, dvipsnames]{color} 
\usepackage{graphicx} 
\usepackage[cmex10]{amsmath} 
\usepackage{amsfonts}
\usepackage{verbatim} 
\usepackage{subfigure}
\usepackage{multirow}
\usepackage{cite}
\usepackage{hhline}
\usepackage{bm}
\usepackage{tabularx,booktabs}
  \newcolumntype{Y}{>{\centering\arraybackslash}X}
\usepackage[bottom]{footmisc}
\usepackage{hyperref}
\usepackage{esvect}

\usepackage{flushend}

\begin{document}
\title{\LARGE \bf Vision- and tactile-based continuous multimodal intention and attention recognition for safer physical human-robot interaction
}
\author{Christopher~Yee~Wong, Lucas Vergez, and Wael~Suleiman
\thanks{C. Y. Wong, L. Vergez, and W. Suleiman are with the Universit\'{e} de Sherbooke, Sherbrooke, Canada (e-mail: \texttt{christopher.wong2}, \texttt{lucas.vergez}, \texttt{wael.suleiman [at] usherbrooke.ca}).}
}

\maketitle
\thispagestyle{empty} 
\pagestyle{empty} 


\begin{abstract}
Employing skin-like tactile sensors on robots enhances both the safety and usability of collaborative robots by adding the capability to detect human contact.
Unfortunately, simple binary tactile sensors alone cannot determine the context of the human contact---whether it is a deliberate interaction or an unintended collision that requires safety manoeuvres.
Many published methods classify discrete interactions using more advanced tactile sensors or by analysing joint torques.
Instead, we propose to augment the intention recognition capabilities of simple binary tactile sensors by adding a robot-mounted camera for human posture analysis.
Different interaction characteristics, including touch location, human pose, and gaze direction, are used to train a supervised machine learning algorithm to classify whether a touch is intentional or not with an F1-score of 86\%. 
We demonstrate that multimodal intention recognition is significantly more accurate than monomodal analyses with the collaborative robot Baxter.
Furthermore, our method can also continuously monitor interactions that fluidly change between intentional or unintentional by gauging the user's attention through gaze.
If a user stops paying attention mid-task, the proposed intention and attention recognition algorithm can activate safety features to prevent unsafe interactions.
We also employ a feature reduction technique that reduces the number of inputs to five to achieve a more generalized low-dimensional classifier.
This simplification both reduces the amount of training data required and improves real-world classification accuracy.
It also renders the method potentially agnostic to the robot and touch sensor architectures while achieving a high degree of task adaptability.
\end{abstract}

\def\abstractname{Note to Practitioners}
\begin{abstract}
Whenever a user interacts physically with a robot, such as in collaborative manufacturing, the robot may respond to unintended touch inputs from the user.
This may be through body collisions or that the user is suddenly distracted and is no longer paying attention to what they are doing. 
We propose an easy-to-implement method to augment safety of physical human-robot collaboration by determining whether the touch from a user is intentional or not through the use of robot-mounted basic touch sensors and computer vision.
The algorithm examines the location of the user's hands relative to the touched sensors in addition to observing where the user is looking.
Machine learning is then used to classify in real-time, with an F1-score of 86\%, whether a touch is intentional or not such that the robot can react accordingly.
The method is particularly applicable in collaborative manufacturing contexts, but can also be applied anywhere where a user physically interacts with a robot. 
We demonstrate the utility of the method in enhancing safety during human-robot collaboration through a simulated collaborative manufacturing scenario with the robot Baxter, but the method can easily be adapted to other systems.
\end{abstract}

\begin{IEEEkeywords} 
Gesture, Posture and Facial Expressions;
Human-Robot Collaboration;
Industrial Robots;
Manufacturing;
Multi-Modal Perception for HRI
\end{IEEEkeywords} 

\section{Introduction and Relevant Work} \label{sec:intro}

\begin{figure}[t] 
\centering
\includegraphics[width = 0.35\textwidth]{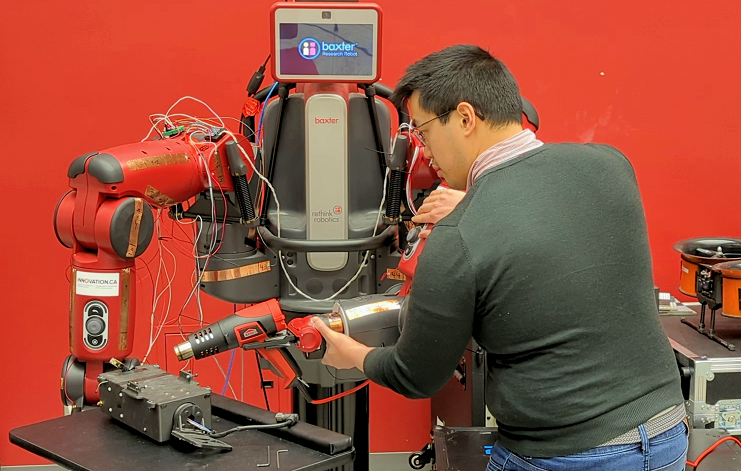}
\caption{A human-robot collaborative scenario where a tool held by a robot is guided by having a user physically manipulate the robot}
\label{fig:cobottoolholding}
\end{figure}

\IEEEPARstart{W}{hen} serial manipulator robots were first introduced in manufacturing, they were separated by cages and walls.
Their large size, heavy arms, and lack of sensors posed great danger to humans if a collision was to happen. 
As manufacturing enters the modern age of Industry 4.0, the trend is not only cageless co-existence of humans and robots in the same space, but humans and robots interacting and collaborating together to complete a task \cite{Raessa2020TASE-HRCAssembly, Liu2021TASE-UnifiedIntentionandLearningforHRC, Armleder2022GChengSkinpHRC} with improved results \cite{Wang2020TASE-VRWeldingHumanIntentionRecognition}.
Sometimes a physical connection exists between the human and the robot either directly through touch or indirectly through an object, leading to what is called physical human-robot interaction (pHRI), as seen in Fig. \ref{fig:cobottoolholding}.

In all cases, safety of the human, the robot, and the environment must all be ensured. 
While many systems rely on establishing safety zones to avoid dangerous collisions between a human and a robot \cite{Rybski2012IROS-WorkcellSafety, Zanchettin2016TASE-HRCSafetyMetrics, Choi2022RCIM-AR-HRCsafety}, these methods are inappropriate for pHRI since pHRI, as the name implies, requires intentional physical proximity and contact.
Therefore, other methods to increase safety are required.
When it comes to physically manipulating a robot, for example in the context of tool guidance, safety in pHRI can be increased either through robot-focused or human-focused methods. 

Robot-focused safety methods focus on improving safety through the hardware and software used on the robot. 
Compliant robotic structural components (e.g., compliant joints or links) \cite{She2020JMD-JointLinkStiffImpactStudy} or compliant skin on soft robots \cite{Goncalves2022RobotSoft-UpperBodyBalloonSkin} provide a hardware-based method for safer interactions.
Compliant control methods, for example impedance and admittance control, are software methods to regulate interaction forces experienced by the human partner through motor torque or force control \cite{Hogan1985-ImpCon1, Ficuciello2015TRO-VariableImpCon-pHRI}. 
An admittance control loop can ensure user safety when the user is physically attached to the robot to perform cooperative tasks \cite{Kim2020ICRA-MOCAMAN}.
Additionally, tactile skin sensors mounted on the body of the robot can further enhance the safety in pHRI by not only detecting unwanted collisions \cite{Papanastasiou2019IJAMT-MultimodalHRCFactoryCase}, but also as a method for enhanced detection of interaction.
Even simple on-off tactile sensors may be used for more complex user intention interpretation to allow intuitive whole-body physical manipulation of humanoid robots \cite{Wong2022THMS-2TKCP}.
More complex tactile sensors incorporating force and proximity detection, among other sensors, may be used in even finer control of the robot \cite{DeanLeon2019ICRA-WholeBodyComplianceSkin} and has been showcased in applications such as dancing \cite{Kobayashi2021JAIM-WholeBodyDancingBoxStep} and kinesthetic teaching \cite{Wrede2013JHRI-KinesTeachUserStudy}.

Human-focused safety methods, comparatively to robot-focused safety methods, involve analyzing the human partner. 
At the most basic level is human pose analysis, which involves examining the physical location and orientation of the user. 
Generalized body location can be used at a larger scale for safety zones as mentioned above \cite{Rybski2012IROS-WorkcellSafety, Choi2022RCIM-AR-HRCsafety}, but the pose of specific limbs may also used for examining human ergonomics \cite{Kim2021RCIM-HRCPowerToolErgonomics} or gesture-based control \cite{deGeaFernandez2017RAS-MultiModalHRC}.
Specifically analysing the head and eyes reveals the notion of gaze. 
Humans look at what they are focused on or what they are interested in \cite{Trick2019IROS-MultimodalIntentionRecog, Xu2020RecognizingUnintentionalTouch, Dias2020WACV-OpenPoseGazeEstimation}, and thus is an excellent predictor of their next move.
While eye gaze provides the best accuracy, determining eye gaze may be difficult if a close view of the face and pupils is not possible. 
Head gaze, on the other hand, is much easier to determine and can usually provide adequate level of information as eye gaze for gross attention \cite{Hollands2002-GazeDirection}, although it is not perfect \cite{Palinko2016IROS-EyeVSHeadTrackingHRC}.
A robot can use gaze to consider human visual–spatial attention to ensure task visibility of its manipulations \cite{Dufour2020RAS-VisualSpatialAttention}.
Physiological indicators can also provide excellent information on the internalized intentions and reactions of the human partner \cite{Hu2022THMS-QuantifyingHumanStatePHRI, Kim2020ICRA-MOCAMAN}. 
One major drawback is the required use of body-mounted sensors, which may be impractical in public areas or robots that interact with many users, but can still provide informative data in smaller laboratory settings. 

While the research cited above examines the different ways to observe humans, robots must be able to interpret and react to this information accordingly.
Intention detection research then focuses on developing methods to translate what a certain input should mean for a robot \cite{Wong2022THMS-2TKCP}.
For example, pHRI studies with humanoid robots and collaborative carrying tasks attempt to translate interaction forces into walking commands \cite{DAgravante2019pHRCCarrying, mohammadi-2019ICRA-ReactiveUBManip}.
Although control algorithms sometimes include disturbance rejection, there is an assumption that the underlying base command is always intentional.

\begin{figure}[t]
\centering
\subfigure[]{
	\includegraphics[height = 3.63 cm]{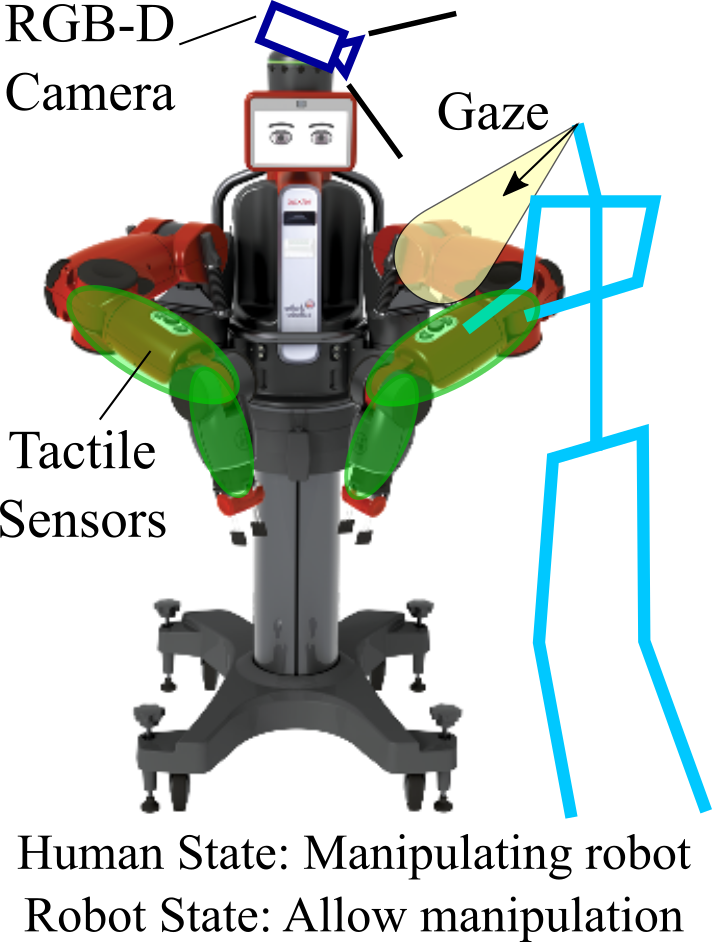}
	\label{fig:interactMANIP}}
\subfigure[]{
	\includegraphics[height = 3.63 cm]{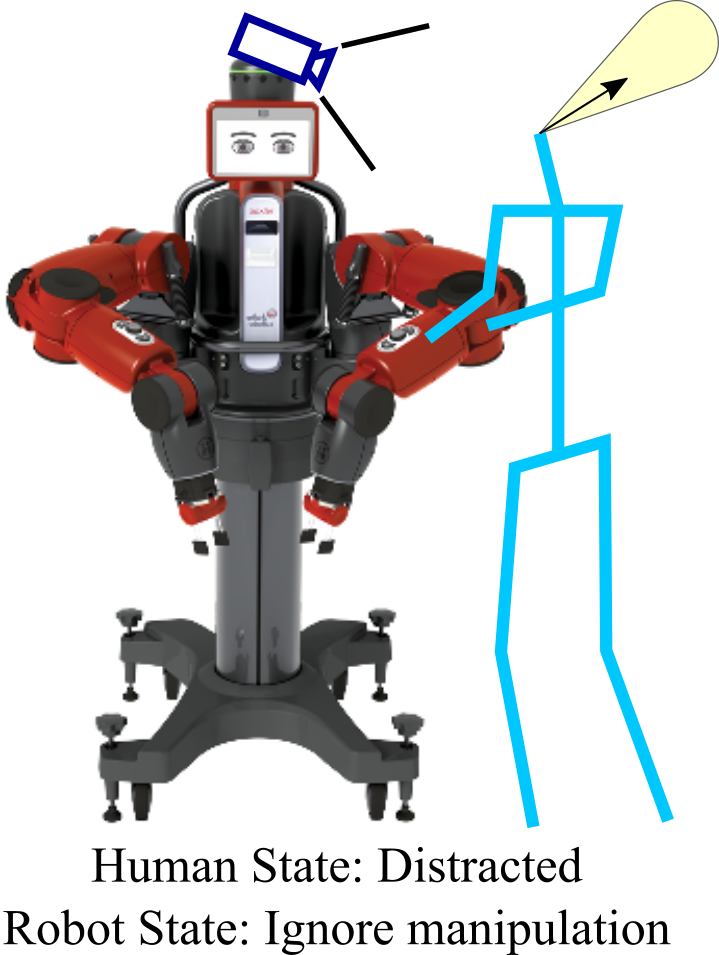}
	\label{fig:interactDISTRACT}}
\subfigure[]{
	\includegraphics[height = 3.63 cm]{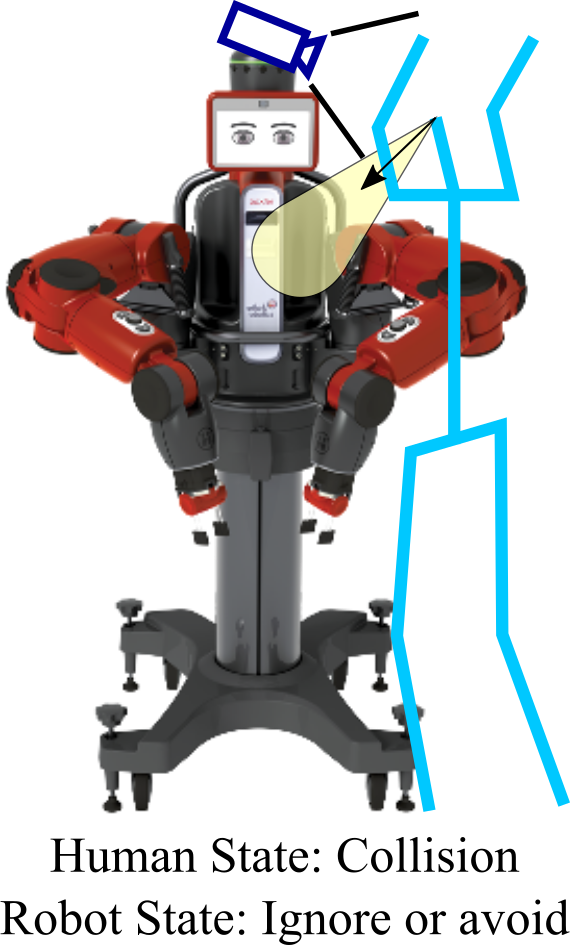}
	\label{fig:interactCOLLIDE}}
\caption{Different interaction cases: b) normal manipulation, c) distracted user, d) unintended collision}
\label{fig:interactCASES}
\end{figure} 

There are a multitude of ways to determine whether an interaction with a robot is intentional or not.
High-resolution tactile skin sensors can form tactile images, so that a convolution neural network can classify whether the contact is a human hand or not \cite{Albini2020IJRR-HandVSBodyContactUsingSkin}.
Such an algorithm could potentially allow a robot to react differently whether a touch is recognized as a hand or non-hand contact.
Deliberate or accidental interactions can also be classified using supervised learning techniques by observing joint torques and interaction wrenches \cite{Lippi2021ICRA-NNTouchRecognition, Briquet-Kerestedjian2019ECC-NNpHRIClassification}.
These techniques are robot-specific and sometimes even require task- and environment-specific data for training, reducing the flexibility of the methods. 
Although only a large tabletop touchscreen is used, \cite{Xu2020RecognizingUnintentionalTouch} showcases a method to filter out unintentional touches by learning user behaviours including eye and head gaze and spatiotemporal touch features.
For example, users usually looked at intentional interaction points and certain unintentional touches tended to cluster more along the edges of the table and be more static (e.g., touches resulting from the elbows resting on the edge of the table).
Combining multiple modalities of intention detection enhances robustness of the intention recognition algorithms \cite{Cooney2012IROS-TouchVisionRecognition, Trick2019IROS-MultimodalIntentionRecog, Wang2021TASE-PredictingHumanIntentionHandover}.



\begin{figure*}[t] 
\centering
\includegraphics[width=0.87\textwidth]{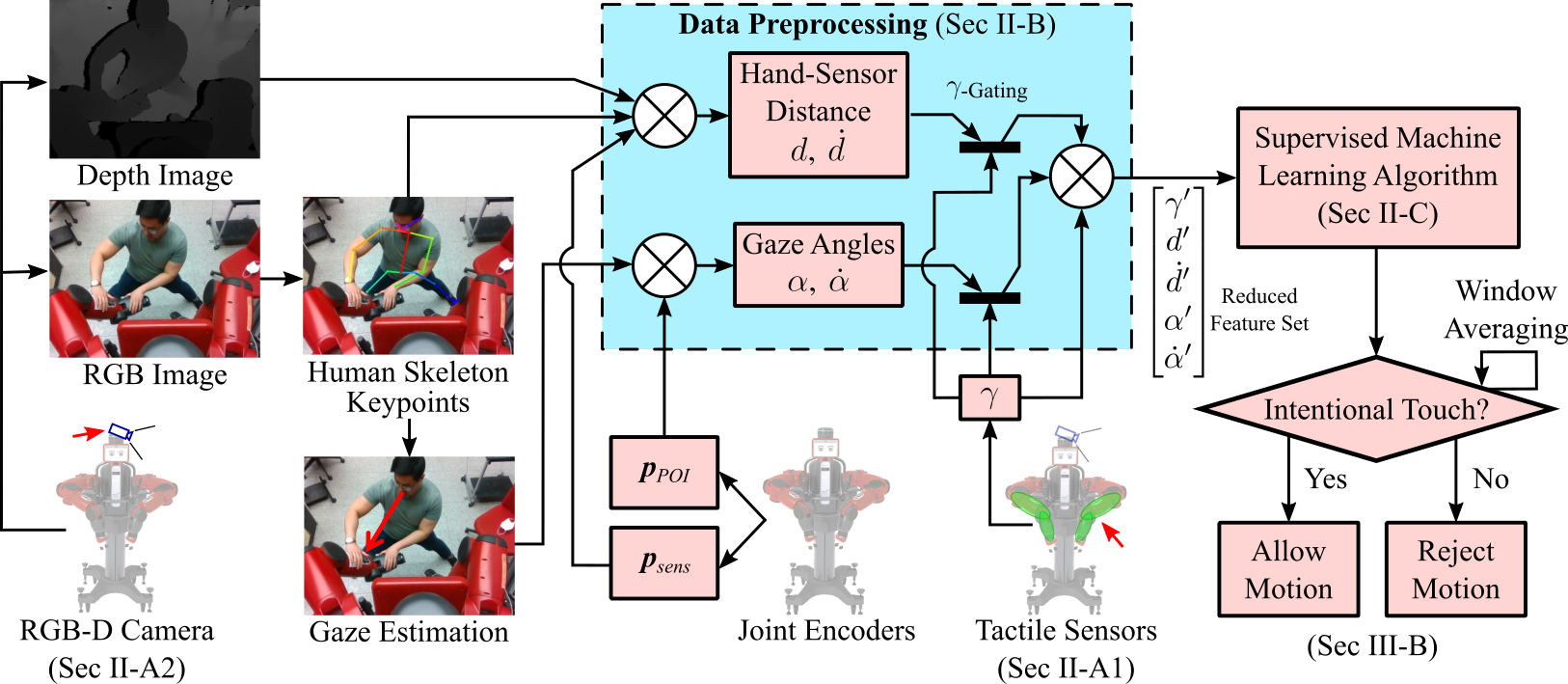}
\caption{Overview of the proposed multimodal intention and attention recognition algorithm. Three main sensors are used: the tactile sensors (Sec. \ref{sec:tactilesensors}), an RGB-D camera (Sec. \ref{sec:rgbdcam}), and the robot joint position encoders. 
The tactile sensors $\gamma$ report a binary state of touched ($\gamma = 1$) or not touched ($\gamma = 0$).
From the RGB image, OpenPose \cite{Cao2021TPAMI-OpenPose} is used to extract the user skeleton keypoints and estimate the gaze direction \cite{Dias2020WACV-OpenPoseGazeEstimation}. 
The keypoints are then projected into Cartesian space using the depth image from the camera.
The joint position encoders are used with forward kinematics to calculate the position of any points of interest $\bm{p}_{POI}$ and the position of the tactile sensors $\bm{p}_{sens}$ in Cartesian space.
These data are then preprocessed (Sec. \ref{sec:datapreprocessing}) and thresholded using the tactile sensor values $\gamma$. 
The feature vector is reduced before becoming the input to the supervised intention classifier (Sec. \ref{sec:MLmodel}), which classifies whether an interaction is intentional or not.
In our demonstration case (Sec \ref{sec:experiments}), the robot will allow or reject a manipulation by varying the stiffness in the impedance controller, according to the intention classification.
Other appropriate actions are possible depending the application.
}
\label{fig:overalldiag}
\end{figure*}

With these concepts in mind, the next step in advancing robot cognition, specifically for safety during pHRI, is to analyse the human to distinguish between intended and unintended inputs and to do so in a continuous fashion.
For example, in a physical manipulation scenario where the robot uses interaction forces as input [Figs. \ref{fig:cobottoolholding} and \ref{fig:interactMANIP}], unintended yet potentially valid inputs could manifest when the user suddenly gets distracted and no longer pays attention to the interaction [Fig. \ref{fig:interactDISTRACT}] or when there is a collision with the user [Fig. \ref{fig:interactCOLLIDE}].
A robot that reacts to unintentional inputs could be disastrous for all and there is a level of uncertainty with having a human in the loop that may change at any time.
While the above classifiers are useful for detecting discrete unintentional interactions during pHRI, there is the need to continuously validate that an intentional interaction remains intentional and to analyse the human posture as a whole, similar to \cite{Xu2020RecognizingUnintentionalTouch} but extended for use with robots.
It is also important to develop methods that are compatible with simple on-off tactile sensors to allow lower-cost systems without joint torque sensors or high-resolution tactile sensors to gain intention recognition capabilities.

In this paper, we propose a human-focused method that continuously performs intention and attention recognition using a multimodal approach in the context of direct physical manipulation of an articulated robot during pHRI.
The proposed approach combines a network of simple binary tactile sensors on the robot links along with several heuristics regarding the human pose extracted markerlessly via computer vision.
The heuristics include the position and speed of the user's hands and head gaze.
This information, after a certain amount of preprocessing, is then used as inputs to a supervised machine learning classifier.
We also provide insights on the utility of combining tactile sensors with computer vision for multimodal analysis and certain heuristics for preprocessing the data. 
In addition, we also provide a feature set reduction method that simplifies the machine learning and training problem for intention classification.
This simplification greatly increases the classification accuracy and renders the method potentially agnostic to the robot architecture and touch sensor layout, and can be applied to sensors that have not been trained on.
These insights will aid future designs and iterations that revolve around human-focused analysis during pHRI.
The algorithm is then applied on the collaborative robot Baxter to demonstrate the utility of the algorithm and its ability to detect fluid changes in attention as well as unintended collisions.
The dataset and classifier model are provided for anyone to use\footnote{\url{https://github.com/chrisywong/multimodaltouchintentiondataset}}.

Additionally, our proposed method is complementary to other methods published in literature for intention detection in pHRI, as we use sufficiently different sensing modalities that do not interfere with each other.
Integrating our method with the others would create an even more encompassing multimodal classification algorithm that would significantly improve overall detection accuracy and redundancy.



The proposed multimodal touch intention recognition algorithm is described in Sec. \ref{sec:alg}, the results of the classifier model training and subsequent experiments are discussed in Sec \ref{sec:results}, and finally we provide concluding comments with future directions in Sec \ref{sec:concl}.


\section{Intention Recognition Algorithm}
\label{sec:alg}

The goal of the proposed multimodal intention recognition algorithm is to determine if a detected touch is intentional and, if it is, to continuously verify whether the user is still paying attention.
The method employs a supervised machine learning algorithm that takes various types of pre-processed data stemming from detected touches using tactile sensors, and human pose and gaze direction from computer vision algorithms.
An overview of the algorithm is shown in Fig. \ref{fig:overalldiag}.
Data pre-processing steps are used to simplify the inputs to the machine learning algorithm. 
The output of the classifier can then be used to enact different control strategies, including safety features used to stop the robot from moving or to automatically turn off the tool being used (e.g. a drill or torch). 
The following assumptions are made:
\begin{enumerate}
	\item The user will only intentionally contact the robot with their hands\footnote{Intentional non-hand contacts will be considered in future work}
	\item Contacts will always activate at least one touch sensor
	\item The user is always paying attention to their gaze location
	\item Only one person is manipulating the robot at a time
	\item The user is always within the frame of the camera
\end{enumerate}

\subsection{Robot Hardware}
\label{sec:hardware}

\begin{figure}[t]
  \begin{minipage}[b]{0.518\linewidth}
    \centering
    \subfigure[]{\includegraphics[width=0.98\linewidth]{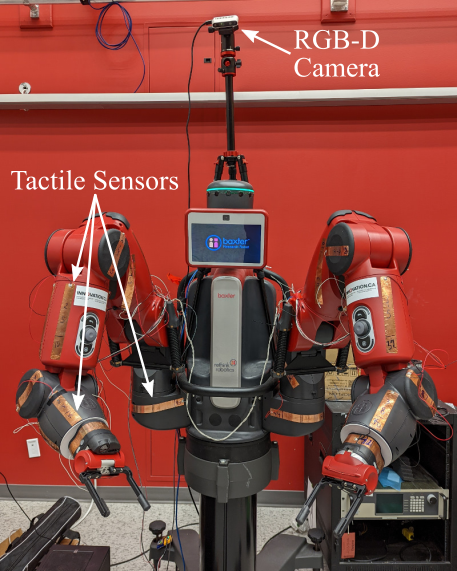} \label{fig:baxter}}
  \end{minipage}%
  \begin{minipage}[b]{0.482\linewidth}
    \centering
    \subfigure[]{\includegraphics[width=0.92\linewidth]{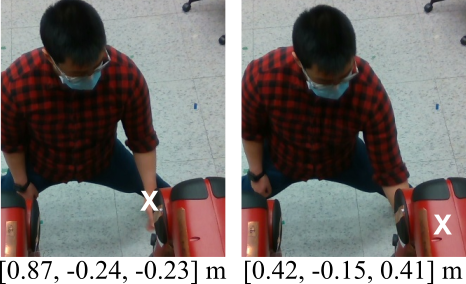} \label{fig:KeypointJumping}}
    \hbox to \linewidth{} 
    \subfigure[]{\includegraphics[width=0.92\linewidth]{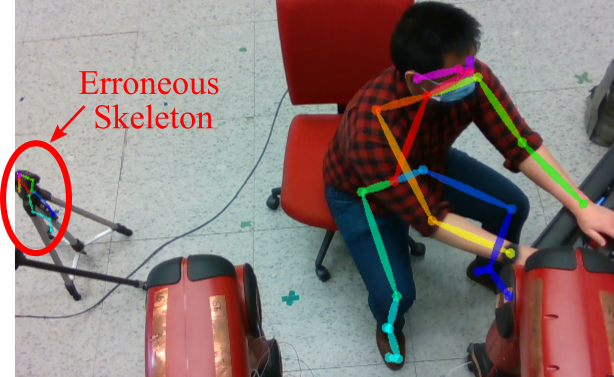} \label{fig:miniskeleton}}
  \end{minipage}
  \caption{a) Baxter robot with custom low-cost tactile sensors and an overhead RGB-D camera; 
	b) keypoint 3D projection error; and c) an erroneous human detected by OpenPose on the left of the image.}
\end{figure}

The collaborative robot Baxter (Rethink Robotics), a fixed-base robot with two arms each with 7 degrees of freedom (DOF) seen in Fig. \ref{fig:baxter}, is used as the experimental platform for demonstration purposes and is augmented with tactile sensors and a RGB-D camera.
All algorithms and robot control are implemented within Robot Operating System (ROS). 



\subsubsection{Tactile Sensors} \label{sec:tactilesensors}
Custom low-cost capacitive touch sensors are mounted on the robot made using strips of conductive copper tape and MPR121 capacitive touch sensor controllers connected to a computer via a FT232H serial-to-USB converter\footnote{\url{https://github.com/chrisywong/ros_ft232h_mpr121}}.
Each arm has 23 tactile sensors (for a total of $n_s = 46$ on the robot) and are simple binary on-off sensors where the sensing values $\bm{\gamma}$ update every 0.4 sec per arm.
The sensors are spread throughout the robot arm.
The bigger limb links are covered by 4 sensors that are set at 90 degrees to each other, while the smaller limb links only have 2 sensors on either side.
The location of the tactile sensors on the limbs are predefined and thus their positions in Cartesian space are known and calculated using forward kinematics.
To note, the number and layout of these sensors are flexible as the proposed classifier only uses an agglomerate relative distance to the hands as input and not multiple absolute positions.
Additionally, the sensors on Baxter, as shown in Fig. \ref{fig:baxter}, are not in an optimized layout and are simply for demonstration purposes only.
Although certain robots, especially collaborative robots, have buttons at the end effector to easily enable free form manipulation, tactile sensors are more flexible in the sense that they allow interactions at places other than the end effector.
This capability is especially important when taking advantage of the redundancy in robots with many degrees of freedom or for robots that do not have such buttons, for example, humanoid robots \cite{Wong2022THMS-2TKCP}.

\subsubsection{Vision System} \label{sec:rgbdcam}
\label{sec:visionsystem}
An Intel Realsense D435 RGB-D camera (colour + depth) is placed approximately 0.85 m above the head of Baxter to provide a slightly overhead view of both the user and the robot. 
The open-source algorithm OpenPose\footnote{\url{https://github.com/CMU-Perceptual-Computing-Lab/openpose}} \cite{Cao2021TPAMI-OpenPose} provides real-time human skeleton keypoint estimation in 2D using 2D images. 
2D skeleton keypoints are then projected into 3D space using depth data from the camera\footnote{\url{https://github.com/ravijo/ros\_openpose}}.
This method has certain drawbacks as OpenPose may estimate a keypoint even if it is occluded, but the simple 3D projection will place the keypoint on the surface of the occlusion rather than behind it, as shown in Fig. \ref{fig:KeypointJumping}.
In this example, the wrist keypoint (white X) is projected on top of the robot limb when the hand is occluded, resulting in a discontinuous jump in position of approximately 0.64 m in the $z$-axis.
This projection error renders it very difficult to determine whether an occluded keypoint is touching the underneath of the robot or not.
The use of tactile sensors in addition to computer vision to detect touch multimodally would partially mitigate this issue by determining whether there is truly contact between the human and the robot for keypoints hidden from view by the robot.

Occasionally, OpenPose will erroneously detect non-existent persons, as shown by the mini skeleton in Fig. \ref{fig:miniskeleton}.
As the erroneous skeletons are typically much smaller than a real person, each skeleton is verified by calculating the limb lengths using the 3D projected keypoints and ensuring that they fall within a nominal range.
Skeletons that do not match these criteria are discarded. 


\subsection{Data Preprocessing} \label{sec:datapreprocessing}

The data obtained by the sensors above are preprocessed using several heuristics prior to being passed to the supervised learning algorithms. 
We define the binary on-off signal of the $j$-th touch sensor $\gamma_j = \{0,1\}$, $j \in 1...n_s$ where $\gamma_j = 1$ for a detected touch.
The features used, relative hand distance $d$, relative hand speed $\dot{d}$, relative gaze angle $\alpha$, and relative gaze speed $\dot{\alpha}$, are scaled between 0 and 1 using their respective maximum values $d_{max}$, $\dot{d}_{max}$, $\alpha_{max}$, and $\dot{\alpha}_{max}$ as seen during training and are constant. 

\subsubsection[Hand distance and speed]{Hand distance $d$ and speed $\dot{d}$}
Assumption 1 states that the only body parts to intentionally contact the robot are the user's hands.
By extension, contacts with any other body part are considered unintentional.
Thus, the only keypoints of interest for proximity analysis are the user's hands.
Although this assumption may be considered limiting, we believe that it encompasses the large majority of use cases. 
The removal of Assumption 1 will be examined in future work.
As physical proximity is a prerequisite for touch, we examine the scalar distance $d_{h,j}$ and the relative speed $\dot{d}_{h,j}$ of hand $h \in \{L,R\}$, where $L$ is the left hand and $R$ is the right hand, to touch sensor $j$ to determine whether it is a hand or a body contact.
The touch sensor signal $\gamma_j$ is used as a gating function (see $\gamma$-gating in Fig. \ref{fig:overalldiag}), as a hand that does not activate a touch sensor is not in contact with the robot and therefore cannot be an intentional touch (Assumption 2).
In other words, if the touch sensor is not activated $\gamma_j = 0$, then $d_{h,j}$ is merely assumed to be the maximum scaled value as distance is inversely proportional to a deliberate touch\footnote{In Sec \ref{sec:modeltraining}, we compare the intention classification performance of learning models using different combinations of features.
In the case where the touch sensors are not used, then touch sensor gating is not used and the distance is always reported.}.
If a touch sensor is activated ($\gamma_j = 1$), then $d_{h,j}$ is calculated as follows:

\begin{equation}
\begin{aligned}
d_{h,j}(t) = 
\begin{cases}
       \frac{1}{d_{max}}\| \bm{p}_{\text{hand},h}(t) - \bm{p}_{\text{sens},j}(t) \| \text{,} &\text{if } \gamma_j = 1 \\
       \text{1} \text{,} &\text{if } \gamma_j = 0 \\
\end{cases}
\end{aligned}
\label{eq:handdist}
\end{equation}

\noindent where $\|\cdot\|$ is the Euclidean distance, and $\bm{p}_{\text{hand},h}$ and $\bm{p}_{\text{sens},j}$ are respectively the positions of the $h$-th hand and $j$-th sensor in 3D space.
The distance is normalized through the division by $d_{max}$.
The purpose of $\gamma$-gating is to ignore irrelevant (i.e. untouched) sensors even if a hand is nearby but not touching.

The relative speed of the hands is potentially an additional factor for determining intention and contact \cite{Zhou2017JoP-ViolentInteractionDetection}.
In the case of non-sliding hand-robot contacts, a hand that is in contact with the robot will have zero relative speed; conversely, a non-zero relative speed would indicate that a hand is not in contact with the robot.
In the case of sliding hand-robot contacts, which are not considered in this paper, the same assumption cannot be made, thus relative speed analysis may still be useful. 
The relative speed $\dot{d}_{h,j}$ of hand $h$ with touch sensor $j$ is computed at each frame as the change in relative distance:

\begin{equation}
\dot{d}_{h,j}(t) = \frac{1}{\dot{d}_{max}}\frac{\lvert d_{h,j}(t) - d_{h,j}(t-\Delta t) \rvert} {\Delta t}
\label{eq:handvelocity}
\end{equation}

\noindent where $\lvert \cdot \rvert$ is the absolute value function and $\Delta t$ is the time step. 
Although the thresholding in (\ref{eq:handdist}) will cause discontinuities in the hand speed calculations, it will only occur for a single time step and will not noticeably impact the intention detection.
While filtering could alleviate these discontinuities, the classifier output is smoothed using a moving window (Sec. \ref{sec:movingwindow}).


\subsubsection[Gaze angle and speed]{Gaze angle $\alpha$ and speed $\dot{\alpha}$}

\begin{figure}[t]
\centering
\includegraphics[width=0.27\textwidth]{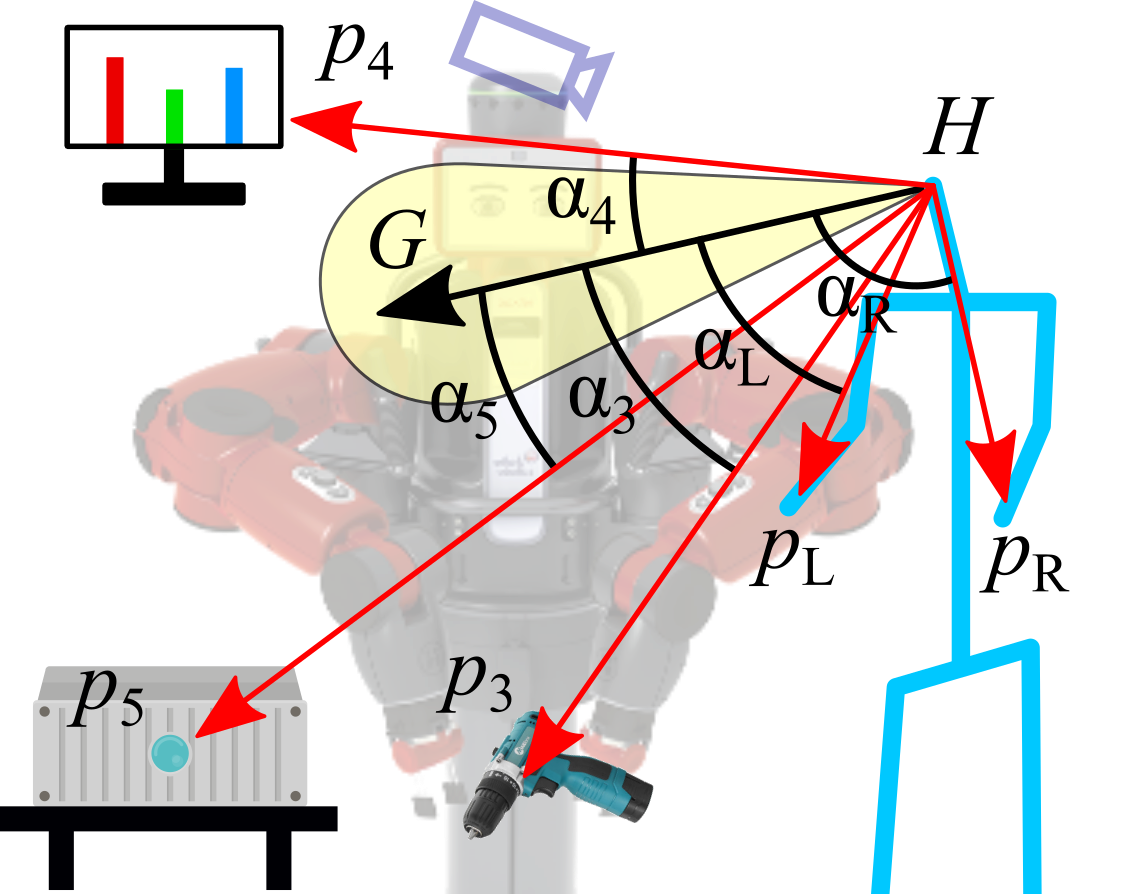}
\caption{Gaze points include the user's hands $p_L$ and $p_R$ and pre-defined points of interest, e.g., the robot end effector or attached tools $p_3$, the computer screen $p_4$, or the workpiece $p_5$.
Relative gaze angle $\alpha_i$ is calculated as the angle between gaze vector $\protect\overrightarrow{HG}$ and position vector of the $i$-th POI $\protect\overrightarrow{Hp_i}$. 
}
\label{fig:gazepoints}
\end{figure}

To estimate the gaze direction, we use the method developed by Dias \textit{et al.} given its accuracy and ease of use as it is based on OpenPose and does not require additional specialized equipment \cite{Dias2020WACV-OpenPoseGazeEstimation}.
A neural network estimates the gaze origin and gaze direction based on relative position and visibility of the various head keypoints.
The gaze direction $\overrightarrow{HG}$ is formed as the vector from the head centroid $H$ to the gaze point $G$, as seen in Fig. \ref{fig:gazepoints}.
The \emph{relative} gaze angle\footnote{For the remainder of the manuscript, we will use the terms \emph{relative gaze angle} and \emph{gaze angle} interchangeably, with the understanding that only angles relative to gaze $\overrightarrow{HG}$ are discussed here.} $\alpha_i$ is then calculated as the angle between $\overrightarrow{HG}$ and the position vector $\overrightarrow{Hp_i}$, i.e., from the head centroid $H$ to the $i$-th point of interest (POI) $p_i$, where $i \in \{POI\}$.
The set of POIs $\{POI\}$ are defined as needed and can include, for example, the hands, the tool held in the robot end effector (EE), the part being worked on, and/or a monitor displaying critical information.
In the current implementation shown in Sec. \ref{sec:results}, the gaze estimation is only implemented in 2D by projecting it onto the camera image plane.
The relative gaze angles are then calculated using these 2D vectors based on pixel location within the camera frame.
In an ideal scenario, using 3D vectors for gaze estimation to generate the relative gaze angle would increase algorithm accuracy, but it is difficult with only a single overhead camera and will be the subject of future work.
 
A special condition exists for the relative gaze angles for the left $\alpha_L$ and right $\alpha_R$ hands: they are only calculated for the hand that is closest to an \emph{activated} sensor.
For example, if multiple touch sensors are activated but the right hand is closer to all the activated touch sensors than the left hand, then only the gaze angle $\alpha_R$ is calculated for the right hand and calculations with the left hand $\alpha_L$ are ignored.
This is to prevent false positives where one hand is touching the robot but the user is looking at their other hand, presumably as an unintentional touch. 
Both $\alpha_L$ and $\alpha_R$ are calculated if they are each respectively closer to at least one touch sensor.
The relative gaze angles for the hands $\alpha_L$ and $\alpha_R$ are calculated as follows:

\begin{equation}
\begin{aligned}
\alpha_{L} = 
\begin{cases}
		\frac{\angle(\overrightarrow{HG},\ \overrightarrow{Hp_{L}})}{\alpha_{max}} \text{,} &\text{if } \gamma_j = 1 \text{ and } d_{L,j} \leq d_{R,j} \\
		\text{1} \text{,} &\text{otherwise}  \\
\end{cases} \\
\alpha_{R} = 
\begin{cases}
		\frac{\angle(\overrightarrow{HG},\ \overrightarrow{Hp_{R}})}{\alpha_{max}} \text{,} &\text{if } \gamma_j = 1 \text{ and } d_{R,j} < d_{L,j} \\
		\text{1} \text{,} &\text{otherwise}  \\
\end{cases}
\end{aligned}
\label{eq:gazeangle}
\end{equation}

\noindent As such, if no touch sensor is activated, then the gaze angles for the hands are set to the scaled maximum of 1 after normalizing by $\alpha_{max}$.
This is similar to (\ref{eq:handdist}) where $\alpha_i$ is inversely proportional to a deliberate touch ($\alpha_i = 0$ indicates that the user is looking directly at the $i$-th POI).
Note that these angles are calculated regardless of what body part is touching the sensors, and that gaze angles for other POIs are not subject to these conditions. 
Gaze angular speed $\dot{\alpha}_i$ is calculated using the backwards difference over a single time step $\Delta t$, similar to (\ref{eq:handvelocity}): 

\begin{equation}
\dot{\alpha}_{i}(t) = \frac{1}{\dot{\alpha}_{max}}\frac{\lvert \alpha_{i}(t) - \alpha_{i}(t-\Delta t) \rvert} {\Delta t}
\label{eq:gazevelocity}
\end{equation}

Hand distance comparison in (\ref{eq:gazeangle}) is akin to assigning whether a touch sensor is touched by the left or right hand.
One potential issue arises when a hand keypoint is occluded and induces a significant depth projection error, as noted in Sec. \ref{sec:visionsystem}.
For small projection errors, such as when touching sensors on the underside of a limb and the keypoint is projected on top of that same limb, then this error is minimal.
If the projection error is large as a result of the hand keypoint being hidden behind multiple limbs, as seen in Fig. \ref{fig:KeypointJumping}, and the non-touching hand is close by, (\ref{eq:gazeangle}) will then mislabel the incorrect hand as closer and provide the incorrect distance and angle calculations.
Additional cameras to view the interaction at different angles could minimize issues caused by occlusions and will be the subject of future work.


\subsubsection{Input Vector and Feature Reduction}
As our robot has $n_s = 46$ touch sensors spread over the whole body, only a small number of these sensors are used at any given time.
A full feature vector that includes every touch sensor signal, hand distances and speeds for each sensor, and gaze angles and speeds for each POI would potentially introduce irrelevant noise to the feature vector.
This is partly mitigated by using $\gamma$-gating.
Furthermore, properly training the learning algorithms would require large amounts of data that includes all conditions for all touch sensors.
Therefore, rather than using a full touch sensor-focused feature vector, we instead choose to examine a reduced aggregated single touch-focused feature set.
This reduced input vector only includes 5 features: $\begin{bmatrix} \gamma' & d' & \dot{d}' & \alpha' & \dot{\alpha}' \end{bmatrix}$ where $\gamma'$ tracks whether \emph{any} touch sensor is activated, and $d'$ and $\alpha'$ are respectively the minimum relative hand distance across all hands/sensors and the minimum gaze angle across all POIs.
In addition, $\dot{d}'$ and $\dot{\alpha}'$ are the hand speed and gaze speed of $d'$ and $\alpha'$:

\begin{equation}
\begin{aligned}
\gamma' &= \max\limits_i(\gamma_i) \\
d' &= \min\limits_{h,j}(d_{h,j}),\quad &\dot{d}' = \frac{1}{\Delta t}\lvert d'(t) - d'(t-\Delta t) \rvert \\
\alpha' &= \min\limits_{i}(\alpha_i),\quad &\dot{\alpha}' = \frac{1}{\Delta t}\lvert \alpha'(t) - \alpha'(t-\Delta t) \rvert
\end{aligned}
\label{eq:reducedfeatureset}
\end{equation}

This reduction of the feature set serves two purposes. 
Firstly, since only the aggregated touch data is used and individual sensor information is removed, the model does not need to be trained for each and every single touch sensor.
This removes the need for comprehensive data to train the system.
Secondly, the algorithm is now potentially robot and touch sensor layout agnostic, as no specific data linked to the system is required.
This would render the algorithm significantly more flexible and easily scalable to other systems.
Although there are insignificant differences in training accuracy between the full and reduced feature sets, we observe significantly better performance in real-world usage with the reduced feature set than with the full feature set.
As an example, when analysing the experiments in Sec. \ref{sec:experiments} as seen in the first plot of Fig. \ref{fig:demo-plot}, the full feature set classifier (light grey) erroneously classifies the intention of the user at multiple time points.
The misclassification between time points D and E-1 is particularly unacceptable for a real use case. 
Therefore, only the reduced feature set is used. 



\subsection{Supervised Machine Learning Algorithm} \label{sec:MLmodel}


The proposed intention and attention recognition algorithm uses a supervised learning classifier where the goal of the algorithm is to return if a touch is intentional or not at each moment.
Given the small number of features used, only a simple classifier is needed for this task. 
The performances of 3 learning models (support vector machines (SVM), neural networks (NN), and $k$-nearest neighbours (kNN)) are compared.
Model instantiation, training, and usage are implemented using \texttt{scikit-learn} \cite{scikit-learn} classes in \texttt{python}.
Unless otherwise stated, the default parameters are used for the various models.
Given that only 5 input features are used and the classification is binary, the neural network architecture consists of a 5-feature input layer and a 1-feature output layer.
Various architectures including different combinations of 1 to 3 hidden layers and 3 to 100 neurons per hidden layer are tested. 
Of the ones tested, a single fully-connected hidden layer with 10 neurons performs the best and is reported here.
The chosen NN architecture is thus $\begin{bmatrix} 5, 10, 1 \end{bmatrix}$.
The \texttt{MLPClassifier} class is used with the \texttt{lbfgs} solver and a maximum of 1000 iterations for training. 
For the kNN model, as we have $n_{samples} = 3002$ training frames, we use the rule of thumb for kNN where $k \approx \sqrt{n_{samples}}$. 
Although we choose $k = 51$ (kNN-51) based on this rule of thumb, the training and testing results indicate that $k = 11$ (kNN-11) has better performance and will be discussed further in Sec. \ref{sec:modeltraining}. 
Euclidean distance is used as the distance metric in kNN and the nearest neighbours are weighted inversely by distance.
As rigorous tuning of the parameters and architectures of the learned models is not the focus of this paper, further optimization could potentially provide more accurate results, perhaps through the \texttt{scikit-learn} function \texttt{RandomizedSearchCV}. 

\subsection{Output Moving Window Smoothing}
\label{sec:movingwindow}

As the gaze estimation can be noisy, especially when the face is not clearly seen or if a facial mask is worn, borderline cases may cause the classifier to erratically switch between intentional and unintentional classifications. 
Thus, to smooth noisy cases, the output of the classifier is averaged over 1 sec using a moving window.
This smoothing improves subjective real-world usage of the intention and attention recognition algorithm.
Furthermore, if the user is switching their gaze between different points of interest, the gaze may pass through a region that would cause the classifier to label the input as unintentional.
A moving window average would then prevent such erroneous classifications during gaze transitions and those caused by discontinuities in sensor or hand distance switching.
Although this windowing also results in a maximum delay of about 0.5 sec (dependent on algorithm frequency) for a classification to change, the successful demonstration in Sec \ref{sec:experiments} in preventing a collision between the tool and the workpiece suggests that this delay is acceptable. 
This parameter could be tuned in the future for better performance. 


\section{Results and Discussion }
\label{sec:results}

\subsection{Model Training and Comparison}
\label{sec:modeltraining}

\begin{figure}[t]
\centering
\subfigure[]{\includegraphics[width=0.22\textwidth]{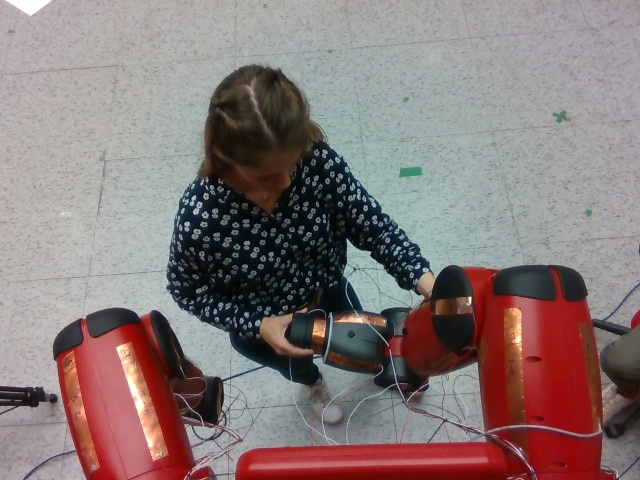}	\label{fig:trainingexample2}}
\subfigure[]{\includegraphics[width=0.22\textwidth]{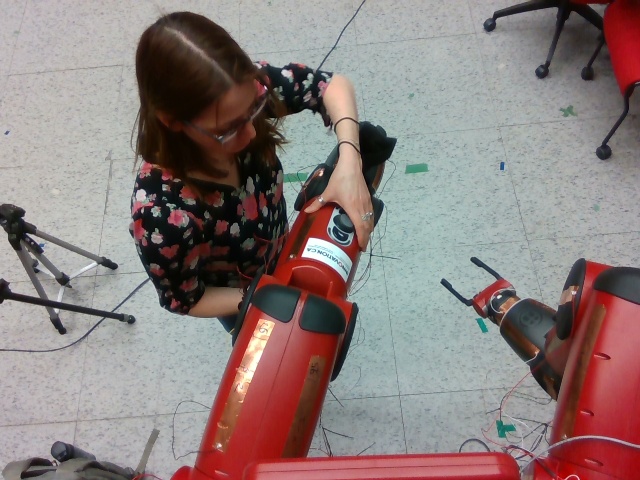}	\label{fig:trainingexample3}}
\subfigure[]{\includegraphics[width=0.22\textwidth]{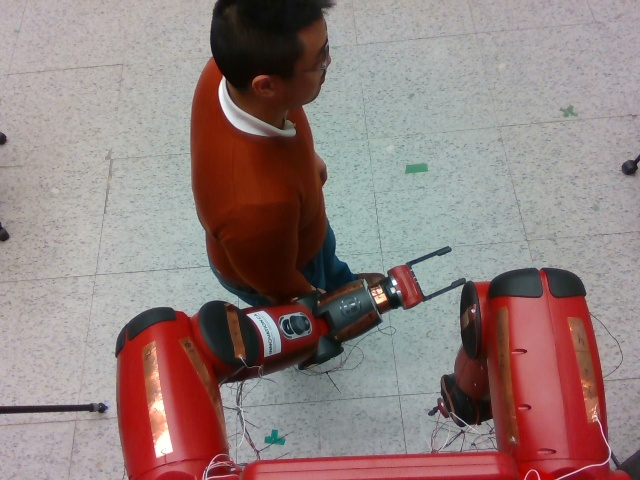}	\label{fig:trainingexample4}}
\subfigure[]{\includegraphics[width=0.22\textwidth]{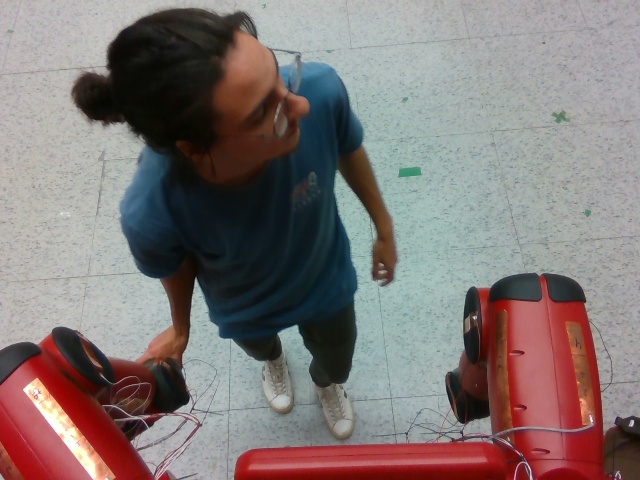}	\label{fig:trainingexample1}}
\caption{Different examples of training data for a-b) intentional and c-d) unintentional interactions.}
\label{fig:trainingexamples}
\end{figure}

\begin{table}[t]
\caption{Model Training Results using 5-Fold Validation $(n_{samples} = 3002)$}
\centering
\setlength{\tabcolsep}{4pt}
\begin{tabular}{c|>{\raggedleft}m{0.25cm} >{\raggedleft}m{0.25cm} >{\raggedleft}m{0.25cm} >{\raggedleft}m{0.25cm} >{\raggedleft}m{0.25cm}|c|cc|cccc}
\toprule
& \multicolumn{5}{c|}{\textbf{Features Used}}                          & \multicolumn{3}{c|}{\textbf{Test Results (\%)}} & \multicolumn{4}{c}{\textbf{Confusion Matrix}} \\
\hline \hline
\parbox[t]{2mm}{\multirow{7}{*}{\rotatebox[origin=c]{90}{\textbf{SVM}\ \ }}} & \footnotesize\textbf{TS} & \footnotesize\textbf{HP} & \footnotesize\textbf{HS} & \footnotesize\textbf{GA} & \footnotesize\textbf{GS} & \textbf{F1} & \textbf{Prec} & \textbf{Recl} & \textbf{TP} & \textbf{TN} & \textbf{FP} & \textbf{FN} \\
\cline{2-13} 
 & $\bullet$ & $\bullet$ & $\bullet$ & $\bullet$ & $\bullet$ & \textbf{80.52} & \textbf{72.71} & 90.21          & 602  & 2082  & 226  & 92   \\
 & $\bullet$ & $\bullet$ & $\bullet$ & $\bullet$ &           & 80.04          & 72.13          & 89.90          & 603  & 2075  & 233  & 91   \\
 & $\bullet$ & $\bullet$ &           & $\bullet$ & $\bullet$ & 80.22          & 72.37          & 89.99          & 605  & 2077  & 231  & 89   \\
 & $\bullet$ & $\bullet$ &           & $\bullet$ &           & 79.96          & 72.06          & 89.82          & 606  & 2073  & 235  & 88   \\
 & $\bullet$ & $\bullet$ &           &           &           & 66.05          & 57.32          & 77.90          & 685  & 1798  & 510  & 9    \\
 & $\bullet$ &           &           &           &           & 64.50          & 55.88          & 76.26          & 694  & 1760  & 548  & 0    \\ 
 &           & $\bullet$ &           &           &           & --             & --             & \textbf{100}   & 0    & 2156  & 0    & 846  \\
\hline \hline
\parbox[t]{2mm}{\multirow{7}{*}{\rotatebox[origin=c]{90}{\textbf{NN} $[5,10,1]$\ \ }}} & \footnotesize\textbf{TS} & \footnotesize\textbf{HP} & \footnotesize\textbf{HS} & \footnotesize\textbf{GA} & \footnotesize\textbf{GS} & \textbf{F1} & \textbf{Prec} & \textbf{Recl} & \textbf{TP} & \textbf{TN} & \textbf{FP} & \textbf{FN} \\
\cline{2-13} 
 & $\bullet$ & $\bullet$ & $\bullet$ & $\bullet$ & $\bullet$ & 82.48          & 75.19          & 91.33          & 606  & 2108  & 200  & 88   \\
 & $\bullet$ & $\bullet$ & $\bullet$ & $\bullet$ &           & 81.79          & 74.36          & 90.86          & 612  & 2097  & 211  & 82   \\
 & $\bullet$ & $\bullet$ &           & $\bullet$ & $\bullet$ & \textbf{82.50} & \textbf{75.25} & 91.29          & 611  & 2107  & 201  & 83   \\
 & $\bullet$ & $\bullet$ &           & $\bullet$ &           & 82.30          & 75.03          & 91.12          & 616  & 2103  & 205  & 78   \\
 & $\bullet$ & $\bullet$ &           &           &           & 74.27          & 65.76          & 85.31          & 651  & 1969  & 339  & 43   \\
 & $\bullet$ &           &           &           &           & 64.50          & 55.88          & 76.26          & 694  & 1760  & 548  & 0    \\ 
 &           & $\bullet$ &           &           &           & --             & --             & \textbf{100}   & 0    & 2156  & 0    & 846  \\
\hline \hline
\parbox[t]{2mm}{\multirow{7}{*}{\rotatebox[origin=c]{90}{\textbf{kNN} ($k$ = 11)\ \ }}} & \footnotesize\textbf{TS} & \footnotesize\textbf{HP} & \footnotesize\textbf{HS} & \footnotesize\textbf{GA} & \footnotesize\textbf{GS} & \textbf{F1} & \textbf{Prec} & \textbf{Recl} & \textbf{TP} & \textbf{TN} & \textbf{FP} & \textbf{FN} \\
\cline{2-13} 
 & $\bullet$ & $\bullet$ & $\bullet$ & $\bullet$ & $\bullet$ & 85.31          & 78.83          & 92.94          & 607  & 2145  & 163   & 87     \\
 & $\bullet$ & $\bullet$ & $\bullet$ & $\bullet$ &           & 84.92          & 78.36          & 92.68          & 612  & 2139  & 169   & 82     \\
 & $\bullet$ & $\bullet$ &           & $\bullet$ & $\bullet$ & \textbf{86.46} & \textbf{80.27} & \textbf{93.67} & 594  & 2162  & 146   & 100   	\\
 & $\bullet$ & $\bullet$ &           & $\bullet$ &           & 84.64          & 77.88          & 92.68          & 595  & 2139  & 169   & 99    	\\
 & $\bullet$ & $\bullet$ &           &           &           & 77.36          & 67.94          & 89.82          & 498  & 2073  & 235   & 196    \\
 & $\bullet$ &           &           &           &           & 66.27          & 54.37          & 84.84          & 417  & 1958  & 350   & 277    \\
 &           & $\bullet$ &           &           &           & 50.42          & 36.54          & 81.31          & 232  & 1753  & 403   & 614    \\
\hline \hline 
\parbox[t]{2mm}{\multirow{7}{*}{\rotatebox[origin=c]{90}{\textbf{kNN} ($k$ = 51)\ \ }}} & \footnotesize\textbf{TS} & \footnotesize\textbf{HP} & \footnotesize\textbf{HS} & \footnotesize\textbf{GA} & \footnotesize\textbf{GS} & \textbf{F1} & \textbf{Prec} & \textbf{Recl} & \textbf{TP} & \textbf{TN} & \textbf{FP} & \textbf{FN} \\
\cline{2-13} 
 & $\bullet$ & $\bullet$ & $\bullet$ & $\bullet$ & $\bullet$ & 83.07          & 76.09          & 91.46          & 627  & 2111  & 197   & 67    \\
 & $\bullet$ & $\bullet$ & $\bullet$ & $\bullet$ &           & 82.49          & 75.39          & 91.07          & 631  & 2102  & 206   & 63    \\
 & $\bullet$ & $\bullet$ &           & $\bullet$ & $\bullet$ & \textbf{84.07} & \textbf{77.29} & \textbf{92.16} & 616  & 2127  & 181   & 78    \\
 & $\bullet$ & $\bullet$ &           & $\bullet$ &           & 82.81          & 75.68          & 91.42          & 616  & 2110  & 198   & 78   \\
 & $\bullet$ & $\bullet$ &           &           &           & 78.09          & 69.02          & 89.90          & 519  & 2075  & 233   & 175   \\
 & $\bullet$ &           &           &           &           & 64.50          & 55.88          & 76.26          & 694  & 1760  & 548   & 0     \\
 &           & $\bullet$ &           &           &           & 54.06          & 39.34          & 86.41          & 190  & 1863  & 293   & 656   \\
\bottomrule
\multicolumn{13}{l}{$\bullet$ = feature is used (TS = touch sensors $\gamma$, HP = hand position $d'$,} \\ 
\multicolumn{13}{l}{HS = hand speed $\dot{d}'$, GA = gaze angle $\alpha'$, GS = gaze speed $\dot{\alpha}'$).} \\ 
\end{tabular}
\label{tab:confusionmatrix}
\end{table}

Forty nine training videos, each 20 seconds long, are taken of five different participants performing various intentional and unintentional manipulations on the robot Baxter, including non-hand contacts with the robot. 
Snapshots of different training samples can be seen in Fig. \ref{fig:trainingexamples}.
Each frame is labelled manually to indicate whether a touch is considered intentional or not.
Twelve videos are discarded for having unusually noisy or corrupted data or the user stepping out of the frame of the camera.
The supervised learning models are then trained on the remaining 37 videos (totalling 3002 frames) using 5-fold validation with frame randomization. 
A detailed breakdown of the training results with different features enabled are tabulated in Table \ref{tab:confusionmatrix}.
The effect of different combinations of features on classification score, averaged across all models, is summarized in Table \ref{tab:performancesummary}.
From a safety standpoint in pHRI, the presence of false positives (FP) is more dangerous than the presence of false negatives (FN) as it can lead to unintended motions. 
Given this imbalanced importance between FP and FN, we include the class specific metrics \emph{precision} (Prec) and \emph{recall} (Recl) and report the combined F1-score, as they are more appropriate measures than basic accuracy.
Generally, it can be clearly seen that training using only the hand position (HP, mean F1-score of 52.24\%) or only the touch sensors (TS, mean F1 of 64.94\%) is inferior to using a multimodal approach and combining the two (mean F1 of 73.94\%).
It is interesting to see that the TS-only SVM, NN, and kNN-51 models simply assume a one-to-one correlation between the touch sensor signal and intention and cannot distinguish unintentional touches.
Similarly, for the HP-only case for the SVM and NN models, all classifications are simply assumed to be negative. 

The inclusion of gaze angle (GA) further improves intention estimation (mean F1 of 82.43\%) by significantly reducing the number of false positives for when a user becomes distracted and looks away, but also slightly increases the number of false negatives when the gaze estimation is noisy and strays far from the actual gaze.
The addition of gaze speed (GS), on average, provides minor improvements to the F1-score (mean F1 of 83.31\%). 
Finally, there is minimal difference when hand speeds (HS) are included (mean F1 of 82.84\%).
While HS may be useful in determining intention in dynamic cases, it is perhaps not as useful for relative motion in kinesthetic manipulation cases as intentional contact with the robot will be redundant with hand distance information.
It may be perhaps useful only in fringe cases where the body is in contact with the robot and the hands are still in motion to confirm that it is in fact not a hand contact.
When looking at the performance of specific models, kNN-11 generally outperforms the other models.

\begin{table}[t]
\caption{Effect of Feature Selection on Scores Across All Models}
\centering
\begin{tabular}{ccccc|c|cc}
\toprule
\multicolumn{5}{c|}{\textbf{Features Used}}                          & \multicolumn{3}{c}{\textbf{Mean} (\%)} \\
\textbf{TS} & \textbf{HP} & \textbf{HS} & \textbf{GA} & \textbf{GS} & \textbf{F1} & \textbf{Prec} & \textbf{Recl}  \\
\hline
$\bullet$ & $\bullet$ & $\bullet$ & $\bullet$ & $\bullet$ & 82.84  & 75.70  & 91.49  \\
$\bullet$ & $\bullet$ & $\bullet$ & $\bullet$ &           & 82.31  & 75.06  & 91.13  \\
$\bullet$ & $\bullet$ &           & $\bullet$ & $\bullet$ & \textbf{83.31}  & \textbf{76.29}  & 91.78  \\
$\bullet$ & $\bullet$ &           & $\bullet$ &           & 82.43  & 75.16  & 91.26  \\
$\bullet$ & $\bullet$ &           &           &           & 73.94  & 65.01  & 85.73  \\
$\bullet$ &           &           &           &           & 64.94  & 55.50  & 78.40  \\
          & $\bullet$ &           &           &           & 52.24  & 37.94  & \textbf{91.93}  \\
\bottomrule
\end{tabular}
\label{tab:performancesummary}
\end{table}

The statistical differences between models are verified by performing McNemar's test ($\alpha = 0.01$) to reject the null hypothesis that the observed difference between two models is due to chance alone.
If rejected, then the observed difference between the two models is considered significant, even if they have similar scores. 
First, we compare between the four different learning models (SVM, NN, kNN-11, and kNN-51) with all five features enabled, i.e., the first line of each model in Table \ref{tab:confusionmatrix}. 
All pairs, except the kNN-11 and kNN-51 pair, reject the null hypothesis and are considered significantly different. 
Second, McNemar's test is applied to compare the seven different combinations of features, as shown in Table \ref{tab:performancesummary}, using only the kNN-11 model. 
All feature-comparison pairs reject the null hypothesis and indicate that the inclusion of different features have statistically significant impact.
%
Given the superior performance of kNN-11 over the other models, we choose kNN-11 as the model to implement in this experiments in Sec \ref{sec:experiments}.
Although kNN-11 and kNN-51 are statistically similar, we continue to choose kNN-11 as it has higher precision in these tests, though more in-depth analysis may yield no difference. 
As a final note, the execution time for the intention recognition algorithm is on average 8.4 ms, which is acceptable for real-time usage. 
The final trained model and training dataset can be found in our online repository\footnote{\url{https://github.com/chrisywong/multimodaltouchintentiondataset}}.

\begin{figure*}[!th]
\centering
\subfigure[]{\includegraphics[width=0.52\textwidth]{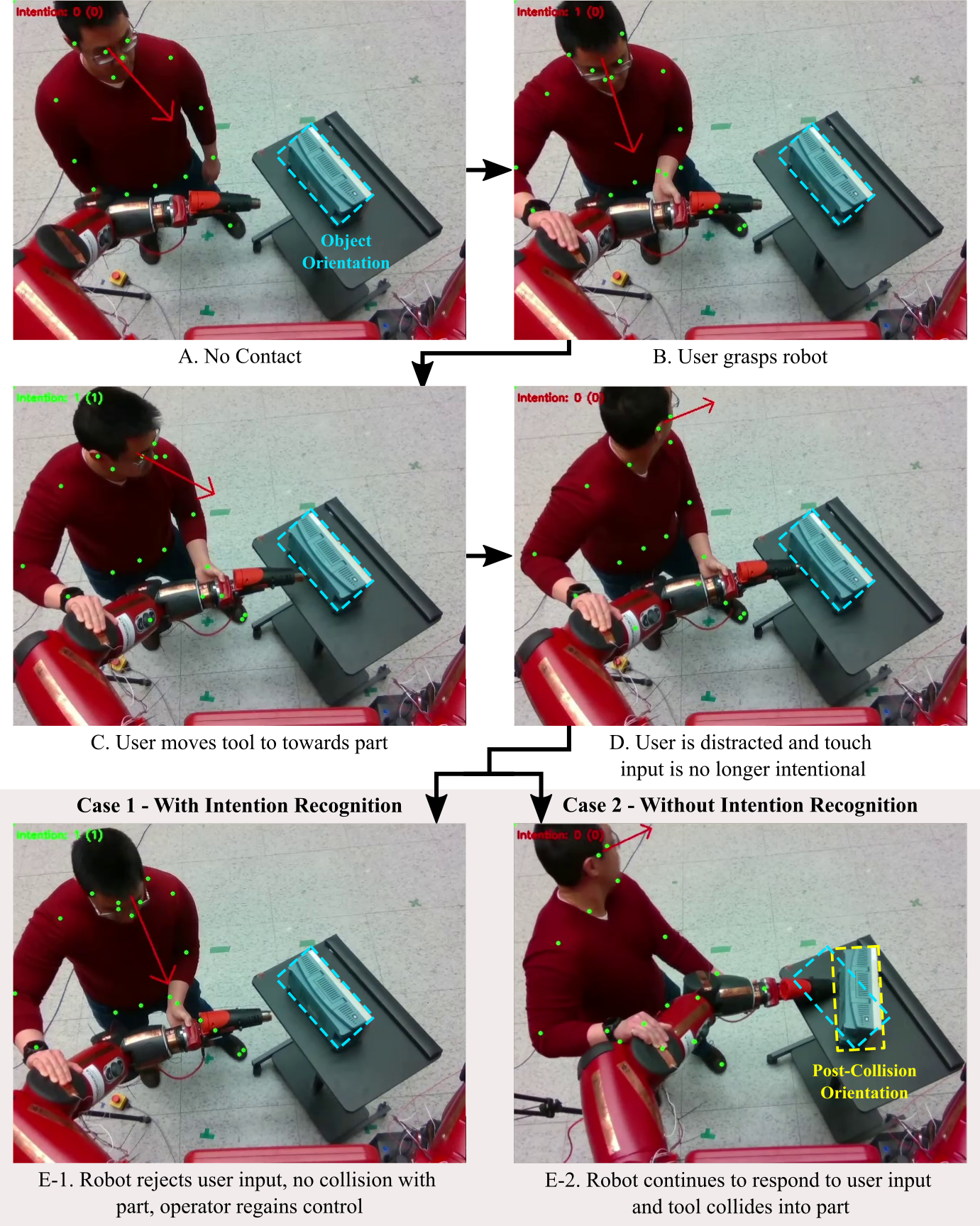}	\label{fig:demo-fig}}
\subfigure[]{\includegraphics[width=0.45\textwidth]{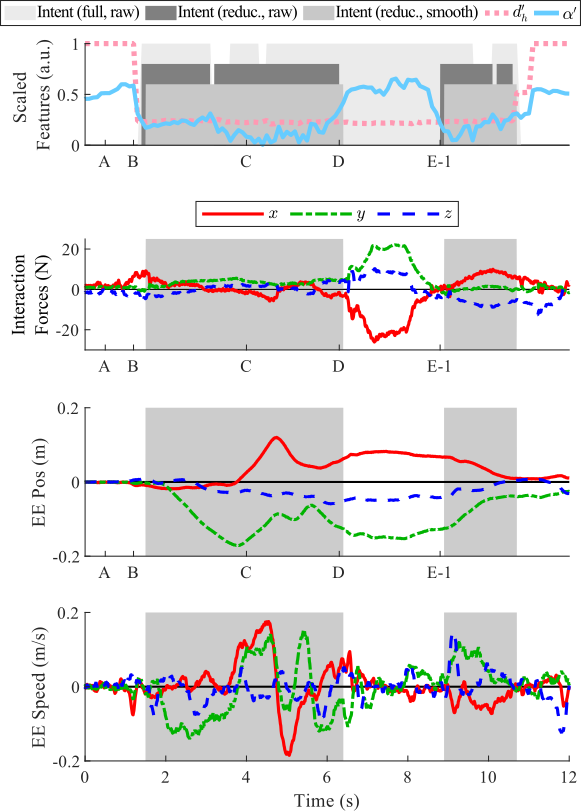} \label{fig:demo-plot}}
\caption{a) A pHRI scenario where a user becomes distracted but continues to manipulate the robot (see supplementary video).
Two cases exist with (Case 1) and without (Case 2) the use of the touch intention recognition-based safety stop. 
Both cases have the same sequence of events up until time point D, where they split into E-1 and E-2 depending on the case. 
The top left label in each frame indicates the intention as detected by the proposed algorithm: green (or 1) for intentional and red (or 0) for unintentional. 
Green dots are the skeleton keypoints and the red arrow indicates the estimated gaze direction. 
b) Plots for Case 1 where a safety stop opposes user force input when the user is no longer paying attention (during the non-shaded areas of the raw/smooth intention using the reduced feature set). 
Comparison with the intention classification using the full feature set shows that the reduced feature set performs significantly better.
} 
\label{fig:demo}
\end{figure*}

Comparatively, other similar methods in literature report F1 scores of 91\% \cite{Xu2020RecognizingUnintentionalTouch} and accuracies in the range of 85 - 94\% \cite{Albini2020IJRR-HandVSBodyContactUsingSkin, Lippi2021ICRA-NNTouchRecognition, Briquet-Kerestedjian2019ECC-NNpHRIClassification}, which are similar to the method proposed here.
We report their accuracies here only for reference as a direct comparison would be inappropriate.
The applications and circumstances of these other methods differ greatly from what is reported in this paper (e.g. the use of a flat static tabletop in \cite{Xu2020RecognizingUnintentionalTouch}, and only performing tactile image classification without an intention component in \cite{Albini2020IJRR-HandVSBodyContactUsingSkin}).
While \cite{Lippi2021ICRA-NNTouchRecognition, Briquet-Kerestedjian2019ECC-NNpHRIClassification} use joint torques to classify touch intention, the methods are dependent on being able to compare actual joint torques to nominal ones, which are only suitable for repetitive tasks and not the free form manipulation performed in this paper. 
That said, it is important to note that our method can instead be used in a complementary fashion with these other methods as our sensor modalities are not incompatible.
For example, the more ideal classifier would expand multi-modal intention detection to encompass all of the reported methods in literature, including ours: body posture analysis through computer vision (our method), high resolution tactile sensors \cite{Albini2020IJRR-HandVSBodyContactUsingSkin} used in a sensor layout-agnostic fashion (our method), joint torque detection \cite{Lippi2021ICRA-NNTouchRecognition, Briquet-Kerestedjian2019ECC-NNpHRIClassification} (for repetitive tasks), accurate eye tracking \cite{Xu2020RecognizingUnintentionalTouch}, etc.
This mix of robot-focused and human-focused methods could then be used to train a classifier with more data modalities to more accurately predict when an interaction is no longer intentional with higher redundancy.



\subsection{Experiments}
\label{sec:experiments}

We demonstrate the intention and attention detection algorithm in a physical human-robot collaboration scenario as shown in Fig. \ref{fig:demo}. 
The included supplemental video\footnote{A link to the supplementary video can also be found in the online repository \url{https://github.com/chrisywong/multimodaltouchintentiondataset}} provides a clearer view of the demonstrations discussed in this section.
In this collaborative scenario, the robot holds a power tool in the end effector while the user guides the robot arm to simulate working on a part.
In potentially dangerous cases like these, it is important to ensure that the tool does not collide with the part or have any extraneous unintended motions.
The proposed intention recognition algorithm can then be used to activate a safety stop that rejects user touch input when the input is considered unintentional.
For demonstration purposes, the robot response to touch input is governed by basic joint impedance control \cite{ott2008cartimptexbook}:

\begin{equation}
\bm{\tau} = -\bm{K}_p(\bm{q} - \bm{q}_d) - \bm{K}_d(\bm{\dot{q}} - \bm{\dot{q}}_d)
\label{eq:impcon1}
\end{equation}

\noindent where $\bm{\tau}$ are the commanded joint torques; $\bm{K}_p$ and $\bm{K}_d$ are the matrices of stiffness and damping coefficients, respectively; $\bm{q}$ and $\bm{q}_d$ are respectively the actual and desired joint positions; and $\bm{\dot{q}}$ and $\bm{\dot{q}}_d$ are respectively the actual and desired joint velocities.
For demonstration purposes here, the safety stop is implemented by having the controller modulate between low stiffness $\bm{K}_p$ for intentional touches and high stiffness for unintentional touches. 
The desired joint positions $\bm{q}_d$ reset to the current joint positions $\bm{q}$ whenever the intention state changes.

It is important to note, however, that the method in which a robot reacts to unintended interactions is arbitrary.
Some examples of intervention methods include restricting the motion of the robot, turning off the tool, triggering an audible alert, or taking no action and simply logging that the user was distracted.
For the purposes of demonstration in this paper, we arbitrarily choose to restrict the motion of the robot when an interaction is deemed unintentional.

In the demonstration shown in Fig. \ref{fig:demo-fig} using the kNN-11 model, we can see the utility of such an intention recognition feature in pHRI scenarios. 
The hand distance $d'$, gaze angle $\alpha'$, corresponding interaction forces, and end effector position and velocity plots are shown in Fig. \ref{fig:demo-plot}.
Beginning with time point A, the user is standing beside the robot with no contact.
The user then reaches out and grabs hold of the robot, making contact at time point B and begins to provide touch input to the robot. 
During normal operation at time point C, there is no interference with the user inputs and the user is able to move the robot end effector to guide the tool along the part.
At time point D, the user is distracted and looks away from what they are doing but continues to push the robot arm. 
In Case 1, where a safety stop is implemented using the intention recognition algorithm, it is determined that the touch input is no longer intentional starting at time point D.
The robot then resists any external forces from the user, as can be seen by the high interaction forces and minimal EE motion between time points D and E-1.
At time point E-1, the user starts paying attention to the robot again and thus regains control of the robot and is able to manipulate the arm without resistance.

Comparatively, the same demonstration is performed again in Case 2 without the intention recognition algorithm in place to reject the user input.
The same sequence of events occurs between time points A-D, as seen in Fig. \ref{fig:demo-fig}.
The difference is that from time point D, the robot continues to respond to the user input even though the user is no longer paying attention and the tool collides into the work piece, as can be seen in the Case 2 at time point E-2 in Fig. \ref{fig:demo-fig}.

In the first plot shown in Fig. \ref{fig:demo-plot}, there are two instances where the average window smoothing is useful.
At $t$ = 3.1 s and again at $t$ = 10.1 s, the user is switching their gaze between two POIs, as can be seen by the small bumps in gaze angle that decrease again quickly. 
At these two instances, the raw output of the classifier is unintentional, but the smoothed output remains intentional.
Without the smoothing, the motion of the arm would have been jerky as the robot would resist motion for a split second. 
As expected, the window smoothing introduces a very short delay but it is still acceptable as the robot motion stopped before colliding with the workpiece. 

\begin{figure}[t]
\centering
\subfigure[]{\includegraphics[width=0.49\textwidth]{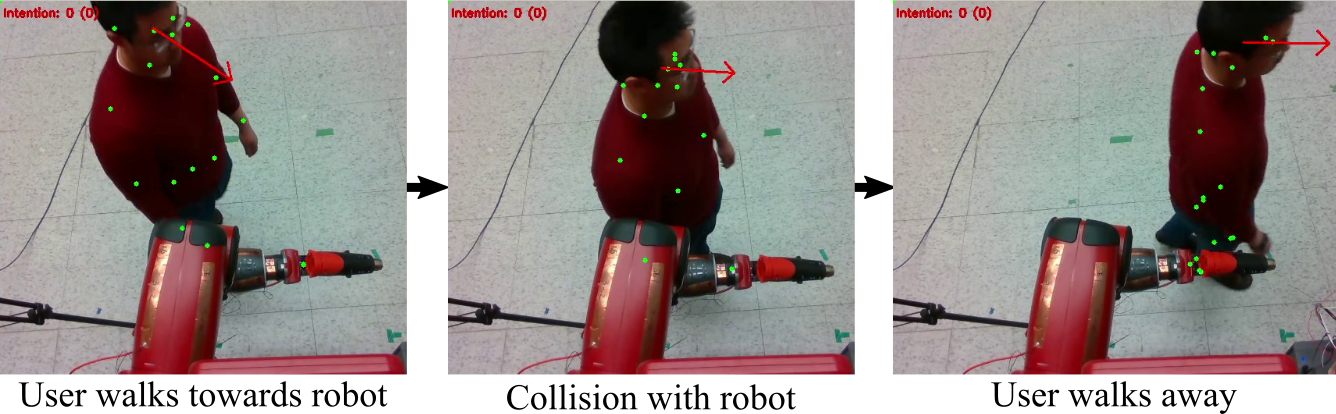}	\label{fig:collision-fig}}
\subfigure[]{\includegraphics[width=0.47\textwidth]{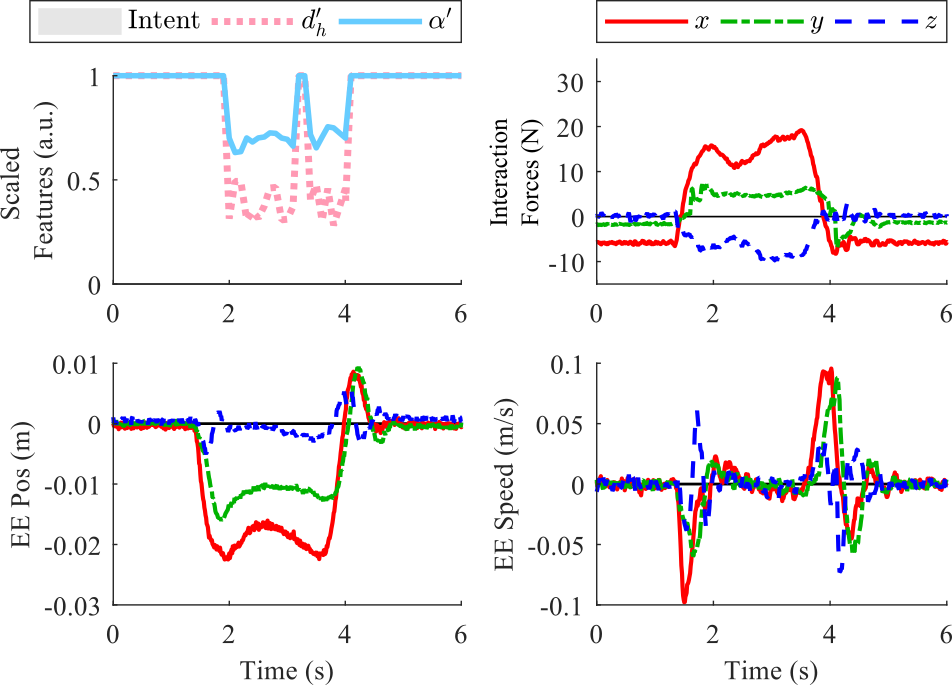} \label{fig:collision-plot}}
\caption{a) A scenario where a user has a body collision with the robot. 
Our algorithm classifies the entire collision sequence as unintentional (no shaded areas) and thus resists any motion resulting from the interaction forces.
b) Plots of certain touch features, interaction forces, and robot motion during the collision scenario.} 
\label{fig:collision}
\end{figure}

In a third experiment, shown in Fig. \ref{fig:collision}, the proposed intention recognition algorithm is also able to properly classify, and subsequently resist, unintentional body collisions with the robot as well. 
During the collision, the robot resists the human contact, as evidenced by the high interaction forces and little movement in the robot end effector position (approximately 0.02 m deviation). 
Note that there are no shaded areas in the plots since there are no moments where the touch is considered intentional. 

\subsection{Task Adaptability}
\label{sec:taskadaptability}

Although the proposed method is demonstrated using these specific examples, we would like to point out several factors that strongly suggest that the proposed method demonstrates a high degree of task adaptability. 
Firstly, the training dataset, as shown in Fig. \ref{fig:trainingexamples}, consists of random interactions with the robot that have no relation to the experimental tasks in Sec. \ref{sec:experiments}.
These interactions simply involve a user performing random touches on the robot and/or moving the robot arm freely with no defined objective, which is clearly different from the experiments above.
Secondly, the only task-related information encoded in the experiment is the position of the work piece to create an additional gaze point of interest, as specified in Fig. \ref{fig:gazepoints}.
Although this information is technically task-specific, the absolute positions of points of interests are not of importance and do not affect the proposed classifier given that only relative distances and gaze angles are used.
Other than the work piece position, no task information is implicitly nor explicitly embedded into the intention and attention recognition algorithm nor in the training data.
Therefore, we believe that these properties strongly support the expectation that the proposed method has a high degree of task adaptability.

%
%
%
%

\section{Conclusion and Future Work} \label{sec:concl}



In this paper, we propose a human-focused multimodal method to distinguish between intentional and unintentional physical interactions with a collaborative robot using touch, user body pose, and head gaze with an F1-score of 86.46\%.
The algorithm is able to detect if a user stops paying attention to their manipulation and, if so, subsequently activate safety interventions to protect the user, the robot and the environment.
The proposed method only requires basic binary on-off tactile sensors and an RGB-D camera to distinguish between intentional and unintentional touches, removing the need for sophisticated and expensive equipment like joint torque sensors or high-resolution tactile sensors.
The method also provides valuable insights for data preprocessing and feature set reduction techniques that increase the accuracy of the supervised machine learning classifier.
Although not explicitly proven, the feature set reduction potentially renders the method agnostic to the robot type and tactile sensor layout, which eliminates the need to train on all tactile sensors.
Furthermore, the architecture of the method suggests a high degree of task adaptability without having to retrain for each different task.
The results also highlight the importance of multimodal analysis in determining intention during pHRI. 

While the proposed method is not an absolute fail-safe as people can lose focus without changing their gaze, it represents another layer of safeguard for improving safety. 
It can also be used to augment other published methods in a complementary fashion, particularly those using interaction wrenches or more complex touch sensors to detect grasp type or exact input force such as those used in \cite{DeanLeon2019ICRA-WholeBodyComplianceSkin, Albini2020IJRR-HandVSBodyContactUsingSkin, Xu2020RecognizingUnintentionalTouch,Lippi2021ICRA-NNTouchRecognition, Briquet-Kerestedjian2019ECC-NNpHRIClassification}.
Future work includes placing an additional camera closer to head level to discern gaze more accurately than the overhead camera currently used and also the possibility of handling interactions involving more than one person. 
More in-depth user studies in real-world scenarios will also be the target of future work, along with testing the algorithms on different systems to explicitly examine the robot agnostic properties of the method.
Other learning models could also be used to include temporal elements of an interaction, e.g., recurrent neural networks (RNN).
The robot could even use this attention recognition to preemptively prepare other protective measures in case the human partner suddenly performs an undesirable action, such as leaving.




\section*{Acknowledgements}
First and foremost, we would like to thank the anonymous reviewers for their insightful comments. 
We would also like to thank Jonathan Mifundu Nzengi for his work on the tactile sensors, and the Fonds de Recherche du Quebec - Nature et technologies (FRQNT) and the Natural Sciences and Engineering Research Council of Canada (NSERC) for supporting in part this research.

\bibliographystyle{IEEEtran}
\bibliography{HumanoidReferences}

\begin{IEEEbiography}[{\includegraphics[width=1in,height=1.25in,clip,keepaspectratio]{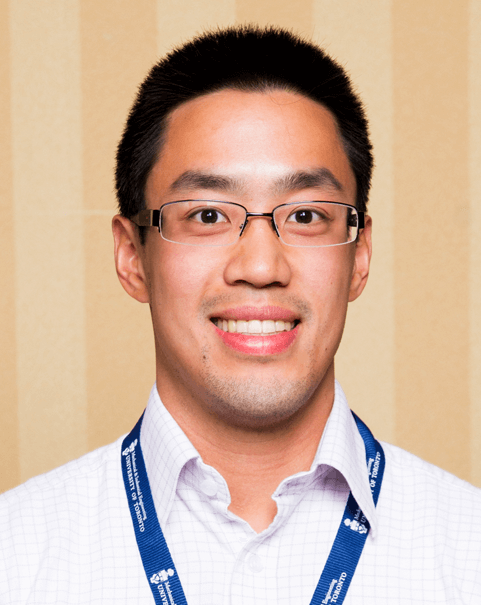}}]{Christopher Yee Wong} (M'15)
 received his B.Eng (2011) and M.Eng. (2014) from McGill University in Montreal, Canada and Ph.D. (2017) in mechanical engineering from University of Toronto in Toronto, Canada. He previously received postdoctoral fellowships for sponsored stays at AIST in Tsukuba, Japan from 2018 to 2019 and at LIRMM in Montpellier, France in 2019. He is currently a postdoctoral fellow at Universit\'{e} de Sherbrooke in Sherbrooke, Canada, where his research focuses on physical human-robot interaction using humanoids, particularly on improving methods for robot cognition with regards to intention detection and human touch.
\end{IEEEbiography}

\begin{IEEEbiography}[{\includegraphics[width=1in,height=1.25in,clip,keepaspectratio]{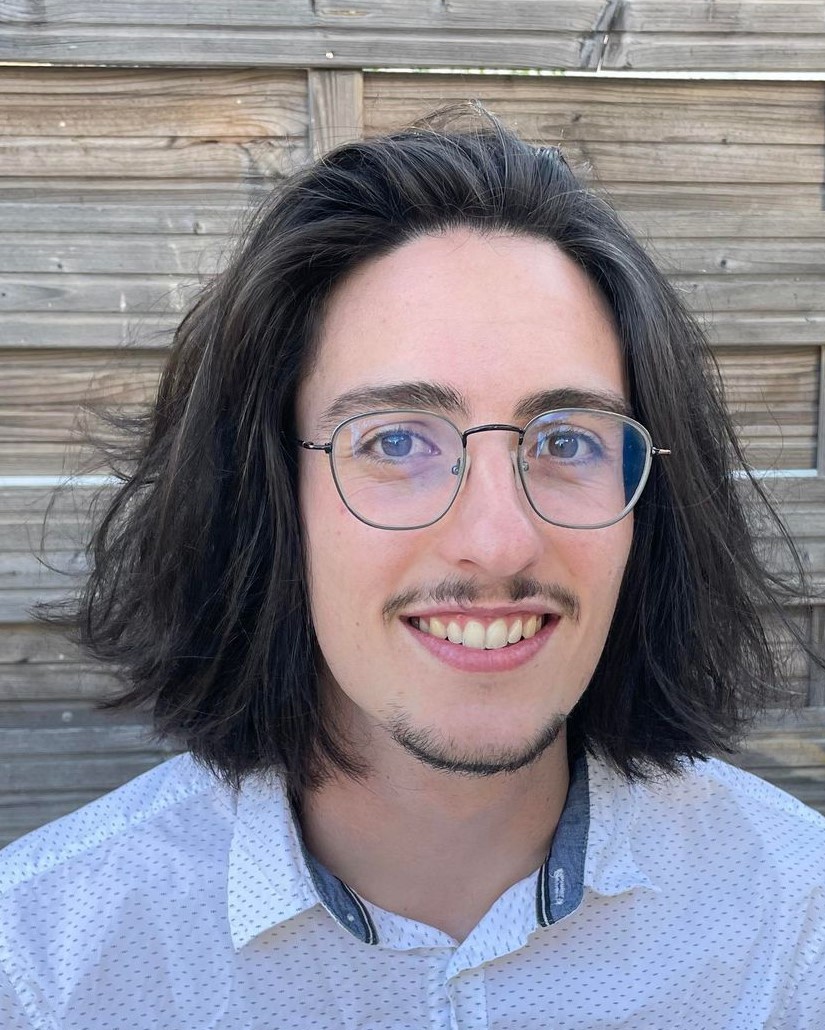}}]{Lucas Vergez} received his M.Eng. (2021) from Arts et M\'{e}tiers Institute of Technology in Aix-en-Provence, France as a graduate in Industrial Engineering and Computer Aided Design. In 2021, he completed a research internship in Computer Vision and Robotics at Universit\'{e} de Sherbrooke in Sherbrooke, Canada and subsequently worked as Virtual Reality Developer in Sherbrooke during 5 months. He is currently a Ph.D. student in Artificial Intelligence and Computer Aided Design at Arts et M\'{e}tiers Institute of Technology, Aix-en-Provence until 2024.
\end{IEEEbiography}

\begin{IEEEbiography}[{\includegraphics[width=1in,height=1.25in,clip,keepaspectratio]{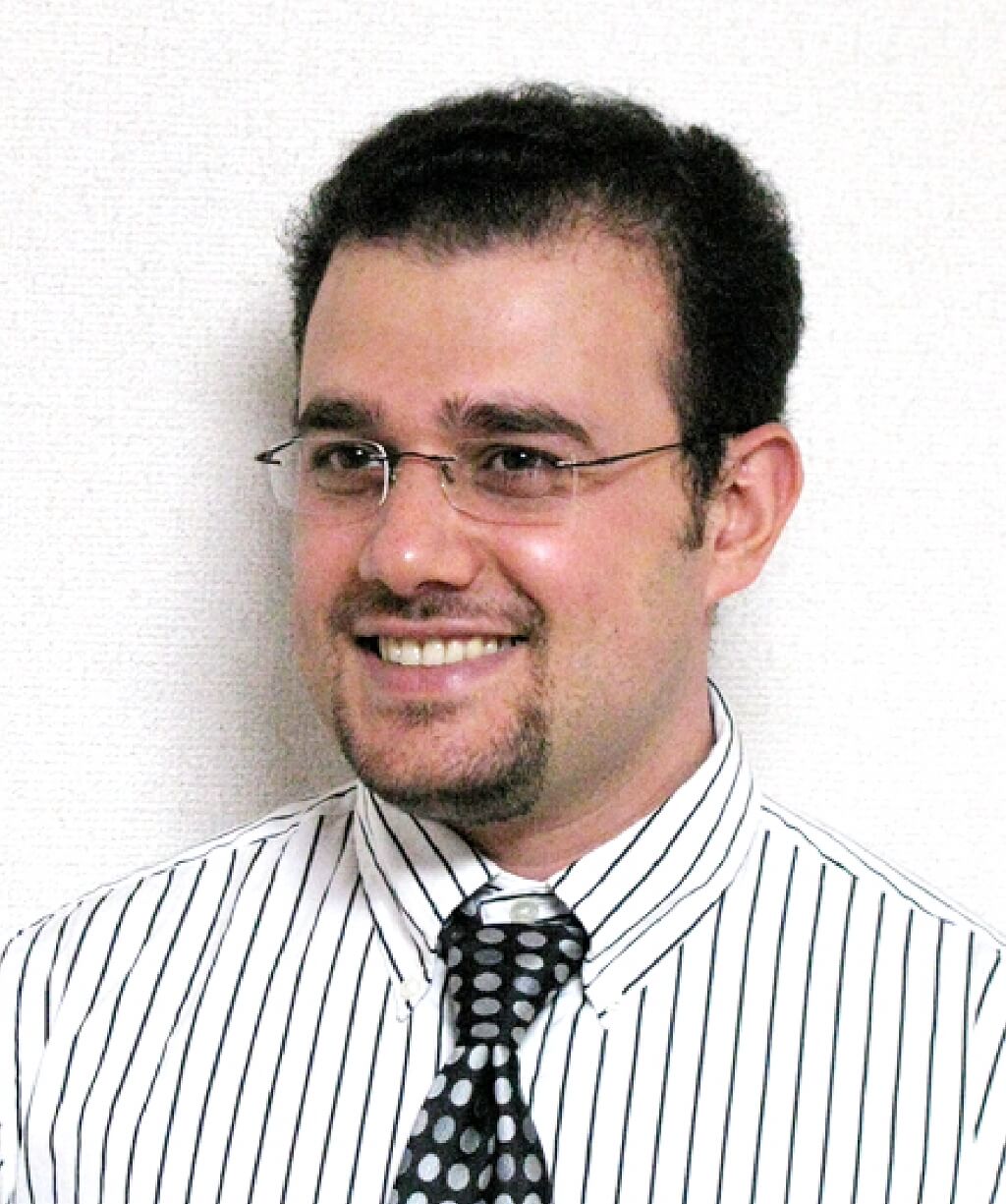}}]{Wael Suleiman} received the Master's and Ph.D. degrees in automatic control from Paul Sabatier University, Toulouse, France in 2004 and 2008, respectively. He has been Postdoctoral researcher at AIST, Tsukuba, Japan from 2008 to 2010, and at Heidelberg University, Germany from 2010 to 2011. He joined University of Sherbrooke, Quebec, Canada, in 2011, and is currently Associate Professor at Electrical and Computer Engineering Department. His research interests include collaborative and humanoid robots, motion planning, nonlinear system identification and control and numerical optimization. 
\end{IEEEbiography}

\end{document}